**Title:** Robotic transcatheter tricuspid valve replacement with hybrid enhanced intelligence: a new paradigm and first-in-vivo study

**Authors:**
Shuangyi Wang,[1,2,3] Haichuan Lin,[1,2] Yiping Xie,[3] Ziqi Wang,[1] Dong Chen,[1,2] Longyue Tan,[1,2] Xilong Hou,[3] Chen Chen,[1] Xiao-Hu Zhou,[1] Shengtao Lin,[4] Fei Pan,[4] Kent Chak-Yu So,[5,6] Zeng-Guang Hou[1,2]*

**Affiliations:**
[1]State Key Laboratory of Multimodal Artificial Intelligence Systems, Institute of Automation, Chinese Academy of Sciences, Beijing 100190, China.
[2]School of Artificial Intelligence, University of Chinese Academy of Sciences, Beijing 100043, China.
[3]Centre for Artificial Intelligence and Robotics, Hong Kong Institute of Science & Innovation, Chinese Academy of Sciences, Hong Kong, China.
[4]Jenscare International Co., Limited, Hong Kong, China.
[5]Division of Cardiology, Department of Medicine and Therapeutics, Faculty of Medicine, The Chinese University of Hong Kong, Hong Kong, China
[6]Division of Cardiology, Department of Medicine and Therapeutics, Prince of Wales Hospital, Hong Kong, China

**Abstract:**
Transcatheter tricuspid valve replacement (TTVR) is the latest treatment for tricuspid regurgitation and is in the early stages of clinical adoption. Intelligent robotic approaches are expected to overcome the challenges of surgical manipulation and widespread dissemination, but systems and protocols with high clinical utility have not yet been reported. In this study, we propose a complete solution that includes a passive stabilizer, robotic drive, detachable delivery catheter and valve manipulation mechanism. Working towards autonomy, a hybrid augmented intelligence approach based on reinforcement learning, Monte Carlo probabilistic maps and human-robot co-piloted control was introduced. Systematic tests in phantom and first-in-vivo animal experiments were performed to verify that the system design met the clinical requirement. Furthermore, the experimental results confirmed the advantages of co-piloted control over conventional master-slave control in terms of time efficiency, control efficiency, autonomy and stability of operation. In conclusion, this study provides a comprehensive pathway for robotic TTVR and, to our knowledge, completes the first animal study that not only successfully demonstrates the application of hybrid enhanced intelligence in interventional robotics, but also provides a solution with high application value for a cutting-edge procedure.

# INTRODUCTION

The prevalence of structural heart disease (SHD) is increasing globally as the population continues to age. Valvular heart disease (VHD) represents the most prevalent form of structural heart disease, with approximately 220 million individuals worldwide estimated to be living with VHD in 2021 alone (*1, 2*). A particular focus on VHD reveals a notable increase in the prevalence of tricuspid regurgitation (TR), which has affected more than 50 million individuals globally in 2021 alone. The advent of prosthetic tricuspid valves has given rise to the emergence of transcatheter tricuspid valve replacement (TTVR) through interventions (*3, 4*), which represents a novel and promising approach in the management of TR. TTVR is a minimally invasive procedure whereby a prosthetic valve is inserted through a delivery catheter via the intercostal or vena cava approach and subsequently released by the surgeon through a series of operations. The superior or inferior vena cava approach is a more minimally invasive technique than the intercostal approach. However, the procedure presents significant challenges with regard to the design of the delivery catheter and the configuration of the valves. To date, only a limited number of valve products have been or are close to being approved for commercial use in TTVR (*5, 6*).

The design of the delivery catheter requires dexterity for precise valve implantation and stiffness for valve release (*7*). It also requires complex mechanical coupling for valve fixation. In the case of prosthetic valves, the majority of designs rely on annular support and clamping of the leaflet structure (*8-10*). However, that might lead to complications such as damage to adjacent structures, unstable fixation of the prosthesis, and even dislocation. Furthermore, the procedure presents evident practical challenges in clinical practice and dissemination. Firstly, the delivery catheter is a sophisticated mechanical device that necessitates external adjustment via a handle outside the body to indirectly alter the position of the end of the device located inside the body. This process is not straightforward and presents a significant challenge. Secondly, the operator is exposed to X-ray radiation for an extended period of time in close proximity, wearing heavy and bulky lead suits, and is at a high risk of developing musculoskeletal and back disorders (*11-14*). In addition, the procedure requires a series of image-based postural adjustments (*15, 16*) to align the end of the delivery catheter with the centre of the tricuspid valve. This process is inherently challenging to achieve with positioning under manual control.

With the development of robotics and artificial intelligence technology, the integration of human operator and machine intelligence is likely to provide an efficient solution to the above problems. Although, to our knowledge, no robotic system for TTVR has been publicly reported, the value of transcatheter interventional robots in cardiovascular disease is well recognized, particularly as exemplified by the vascular interventional robots that have been used to assist with coronary stenting over the past decade (*17-24*). There is a general consensus that robotic assistance can effectively reducing radiation exposure, improving clinical workflow, accelerating the learning curve, and increasing the success rate of the procedure (*25-29*). In 2012, the CorPath 200 System became the first robotic system for coronary intervention to receive U.S. Food and Drug Administration (FDA) approval (*30*). At the present time, robotic-assisted technology for percutaneous coronary intervention (PCI) is the more mature technology in this field. In contrast, research on assisted robots for intracardiac interventional procedures for valve replacement has been limited, largely due to the fact that prosthetic valves and transcatheter valve replacement have only achieved significant clinical breakthroughs just in recent years.

With the clinical advancement, robotic systems for valvular interventions became an emerging hotspot in the last five years, although the number of relevant studies is still limited. The current major work focuses on the robotization of the mitral valve delivery catheter to enable automated and standardized catheter insertion and mitral valve repair. For robotic design, work that has been reported includes robotic add-on drives based on existing commercial delivery catheter (*31, 32*), as well as complete designs from catheter structures to robotic actuation (*33-36*). For motion control, work has been done on modelling and task space control for complex continuum-driven structures within the delivery catheter (*37-39*), as well as on the prospective exploration of ultrasound image-guided catheter control (*40*) and the end-effector's pose estimation under digital subtraction angiography (DSA) (*41*), although the image-based closed-loop control is still a distance from real clinical use. In addition, some of the attempts to address autonomous operations have focused mainly on path planning, such as related work models the surgical approach and constructs a navigation route map based on an inverse reinforcement learning algorithm (*42*). Meanwhile, shape sensing and localization of catheters has been explored, with a focus on the use of electromagnetic and Fiber Brag Grating sensors (*43-46*). The above work on sensing is not specific to valvular interventions and its focus is more relevant to endovascular interventions.

Although the above work demonstrates promising applications for robot-assisted valve interventions, related work is still in the lab testing phase. There is a considerable gap between the exploration of intelligence and autonomy and its practical use in the clinic. Exactly how these intelligent approaches address real-world clinical challenges and what advantages they have over master-slave teleoperation, an approach that is already well productized in vascular interventional robotic systems, has yet to be effectively answered. Considering that the related TTVR products are generally not yet in clinical use, we have developed the concept of robot-assist TTVR system in parallel with the design of new interventional instruments, so as to accelerate the learning curve of the procedure through stable operation and structured, staged intelligent assistance, and to reduce the workload and radiation dose of the surgeon. Unlike many medical robots that were born years after the emergence of manually operated instruments and procedures, we expect that robot-assisted systems for TTVR will be introduced to the clinic at the same time, thus playing a truly important role in the promotion of new clinical procedures.

We aim to propose a complete robotic solution for TTVR, including the prosthetic valve, disposable delivery catheter, and add-on robot drive system. These elements are integrated through a hierarchical, modular design, with a combination of active and passive degrees of freedom and a separation of instrumentation and drive. This solution has real clinical value with a novel interface design that allows the operator to switch between manual and robotic operation while meeting the requirements for sterilization. To facilitate this new approach, we want the robotic system to also be a procedural learning system. To achieve this, we propose an autonomous planning strategy based on digital twins and reinforcement learning, introduce Monte Carlo-based probabilistic maps to cope with the uncertainty of device-tissue interactions and the challenges of image-based localization, and thus enable hybrid enhanced intelligence through human-robot co-piloted control to make full use of the robot's autonomy and the human operator's efficient sensory and cognitive abilities.

In this study, we have conducted a series of phantom tests to demonstrate the effectiveness of the above systems and methods towards autonomy, based on which we have successfully completed in vivo testing of the world's first robot-assisted surgery for TTVR to our knowledge, demonstrating the clinical applicability of the innovative system. The following sections will describe the system in terms of robot design and implementation of hybrid enhanced intelligent methods and report the results of simulation experiments, in-vitro experiments and world's first in-vivo animal testing.

## RESULTS
### Standardized TTVR procedure with robotic cooperation

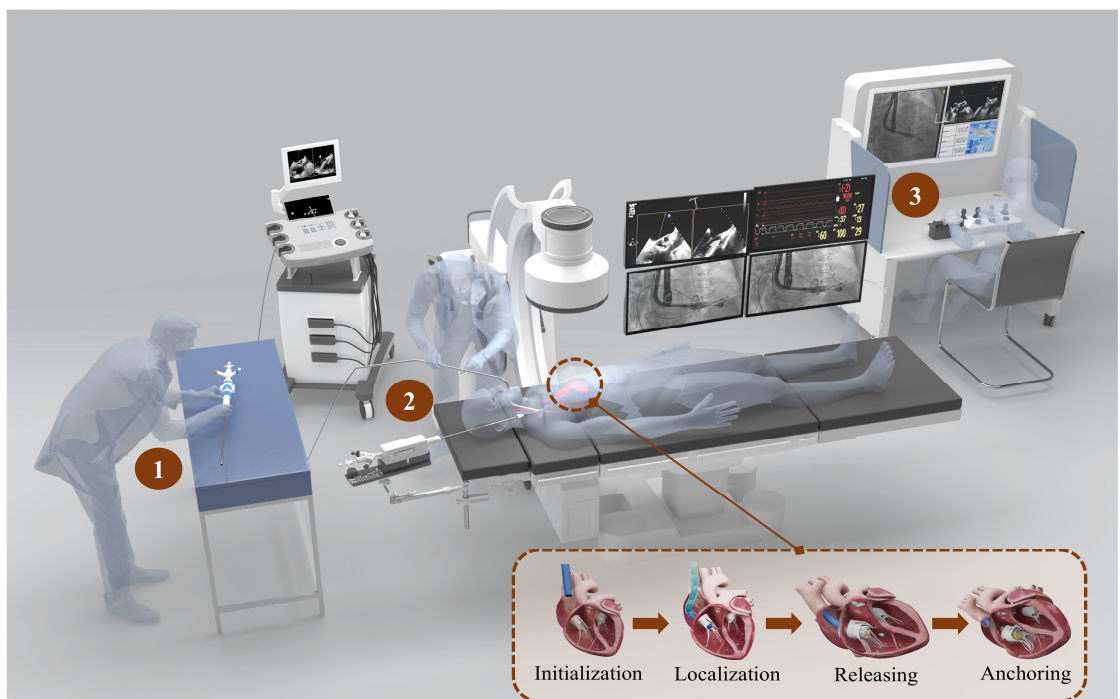

**Fig. 1. Scenario illustration and surgical principle of TTVR procedure.** Label 1 shows the valve being loaded into the delivery catheter prior to surgery, label 2 shows the delivery catheter being introduced into the heart and connected to the robot at the beginning of the surgery, and label 3 shows remote control by the operator through the control console during the main phase of the surgery. The main phases include initialization, localization, releasing and anchoring, and the figure in the dotted line box shows the robot's distal end in the heart at each phase in turn.

TTVR via vena cava access is a novel procedure that has not been widely adopted in the clinic and we aim to standardize the procedure using robotics and hybrid enhanced intelligence. Fig.1 illustrates a scenario for the use of this robotic procedure. A customized robotic system is placed at the bedside to manipulate the delivery catheter. The main operator first installs the valve into the catheter, then assembles the robot and establishes a transjugular access, and installs the delivery catheter onto the robotic drive after manually placing the delivery catheter from the superior vena cava into the atrium, which is referred to as the initialization phase of the procedure. In coordination with the main surgeon there is another sonographer controlling the transesophageal echocardiography (TEE) to provide guidance along with the DSA. Our previous series of work has demonstrated the automated pathways of TEE (*47-53*).

Subsequently, the procedure is robotically assisted, and the operator moves to the main control console which can be deployed away from the X-ray machine or to the control room to minimize or avoid radiation. The operator will perform the two core steps of localization and releasing, where the localization step requires control of the coupled multi-degree-of-freedom motion of the delivery catheter to align with the tricuspid annulus, which requires the operator to have a deep understanding of both the anatomical space and the instrumentation and is the main problem that the hybrid enhanced intelligence approach proposed in this paper seeks to address. The releasing step can be structured into several single degree-of-freedom motions, and the operator is guided through them sequentially by the control software. Finally, the operator will manually perform the one-time action of anchoring by needle sticking, thus completing the entire procedure. An animated illustration of the TTVR procedure can be found in movie S1.

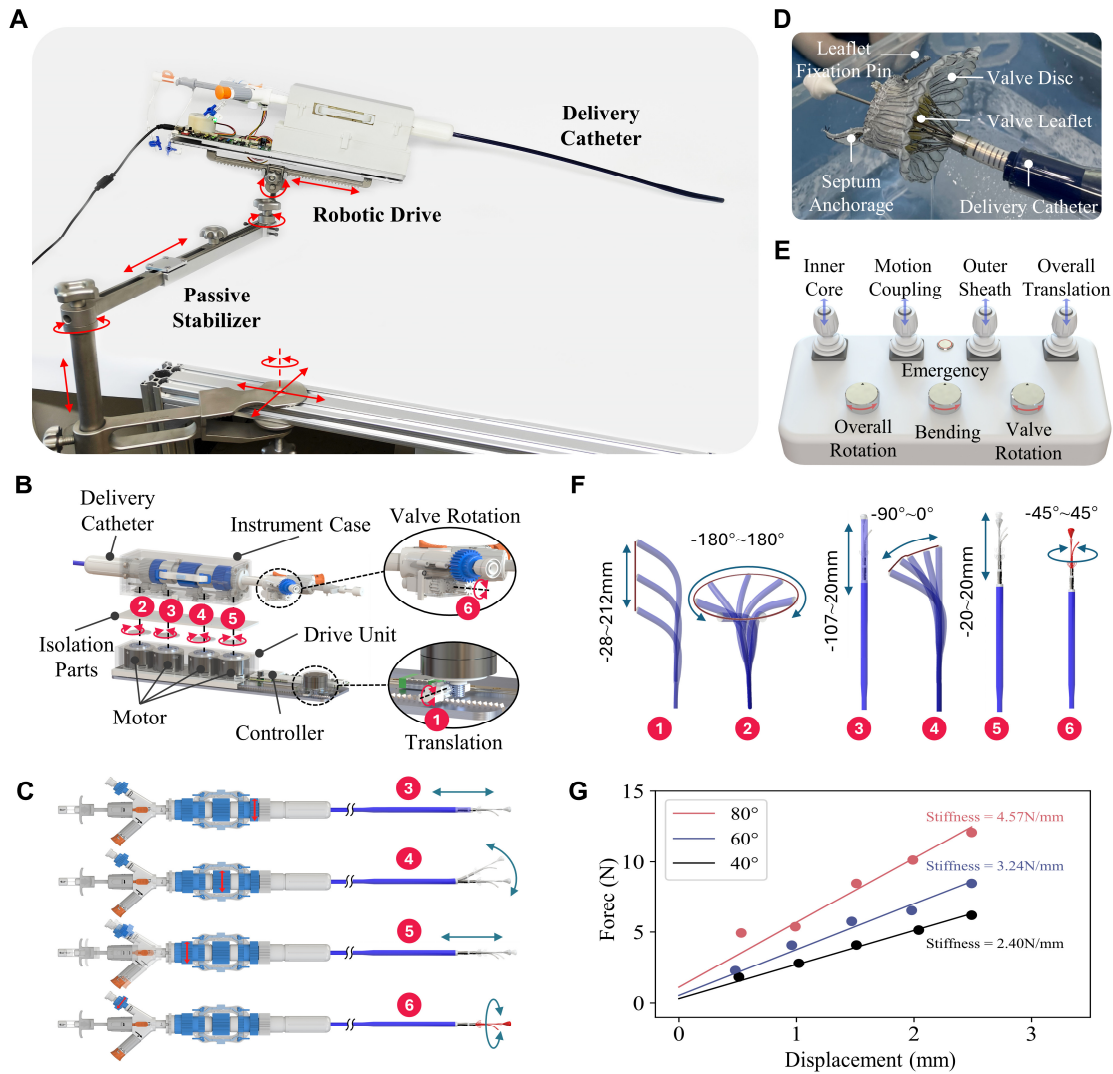

**Fig. 2. Design and function of the robot drive and consumables.** (**A**) Complete demonstration of the robotic system including passive stabilizer, robotic drive and delivery catheter. (**B**) Demonstration of the design principle of the robot's drive, which adopts a design that separates the drive unit from the instrument box and delivery catheter and achieves flexible coupling through a series of mechanical interfaces and transmissions (Labels 1-6 correspond respectively to the overall forward/backward movement, overall rotation of the delivery catheter around the long axis, the advancement/retraction of the outer sheath, the bending of the bending tube, the advancement/retraction of the inner core tube, and the rotation of the gripping jaw inside the inner core tube connected to the valve). (**C**) The structure of the delivery catheter and the available degrees of freedom, including control of the outer sheath, bending tube, inner core tube and gripping jaw. (**D**) Demonstration of the valve structure comprising a valve disc, a valve leaflet, a septum anchorage pin, and two leaflet fixation pins (only one can be seen in the photo angle). (**E**) Demonstration of the master control console, which contains four operating joysticks and three knobs to control the different movements. (**F**) Demonstration of the reachable workspace of each degree of freedom of the delivery catheter. (**G**) Demonstration of the axial stiffness of the end of the delivery catheter under different bending angles.

**Design of the robotic drive and consumables**

The complete structure of the proposed robot is shown in Fig. 2(A). The system consists of three parts: stabilizer, robotic drive and delivery catheter. The stabilizer has nine manually adjustable degrees of freedom, including five prismatic joints and four revolute joints, which are used to adjust the initial position of the delivery system into the human body and to avoid the interference with other objects. As shown in Fig. 2(B), the modularly designed robot consists of a drive unit, isolation parts, instrument box and delivery catheter. A central aspect of the design is sterilization. The isolation parts transmit the power of the drive unit to the delivery catheter and avoid direct contact between them. The above components of the isolation parts are sterilized and in practice a sterile cover could also be fitted between the isolation parts and the drive unit. Another core aspect of the design is the separation of instrument and actuation, which means that the operator can either control the delivery catheter via the robotic drive or remove the robotic drive for manual manipulation. This is achieved by a series of bevel gears for transmitting the power of the drive unit to the delivery catheter coaxial handwheel. The modular design concept is further illustrated in movie S2.

The robotic drive actively controls the six degrees of freedom of the delivery catheter. Motion labels 1 and 2 in Fig. 2(B) are the overall forward/backward movement along the long axis and overall rotation of the delivery catheter about the long axis. The former is realized by a combination of worm gear and rack-pinion and provides a self-locking function, while the latter is realized by a bevel gear transmission. Fig. 2(B) and Fig. 2(C) have a one-to-one correspondence of motion labels 3, 4, 5, and 6, which are all realized by coupling the internal mechanism of the delivery catheter with the gear-based robot drive mechanism. They are, in order, the advancement/retraction of the outer sheath, the bending of the bending tube, the advancement/retraction of the inner core tube, and the rotation of the gripping jaw inside the inner core tube connected to the valve. The bending tube is a double-layer hollow tube structure with the ends welded together and asymmetric grooves cut out on the one side to achieve the bending movement when the concentric tubes are pulled against each other.

The delivery catheter is loaded with a prosthetic valve in the distal end. The structure, shown in Fig. 2(D), consists of a valve disc, a valve leaflet, a septum anchorage pin, and two leaflet fixation pins. The valve leaflet is fixed onto the valve disc, the leaflet fixation pins is used to hook the tricuspid valve leaflet, and the septum anchorage pin is connected and fixed to the interventricular septum. For the control of the robot, the main console, as shown in Fig. 2(E) consists of four operating joysticks and three knobs, where the joysticks are used to control the translational motions, such as overall delivery, linear motion of outer sheath, linear motion of inner core tube, and coupling motion of the inner core tube and outer sheath. The knobs are used to control the rotational motions, such as overall rotation, bending, and gripping jaw rotation. Speed control with linear variations is utilized for fine motion adjustments. Fig. 2(F) shows the theoretical workspace for each active degree of freedom based on its internal mechanism characteristics, combined with transmission and kinematic analyses. Fig. 2(G) illustrates the axial stiffness of the end of the delivery catheter using a force sensor (6-axis, M3552B, Sunrise Instruments, Shanghai, China) and a digital caliper-based displacement measurement unit under different bending angles, where it can be seen that the delivery catheter has a wide range of bending capacity with sufficient stiffness so that external perturbations in the intracardiac environment do not result in a loss of distal end position.

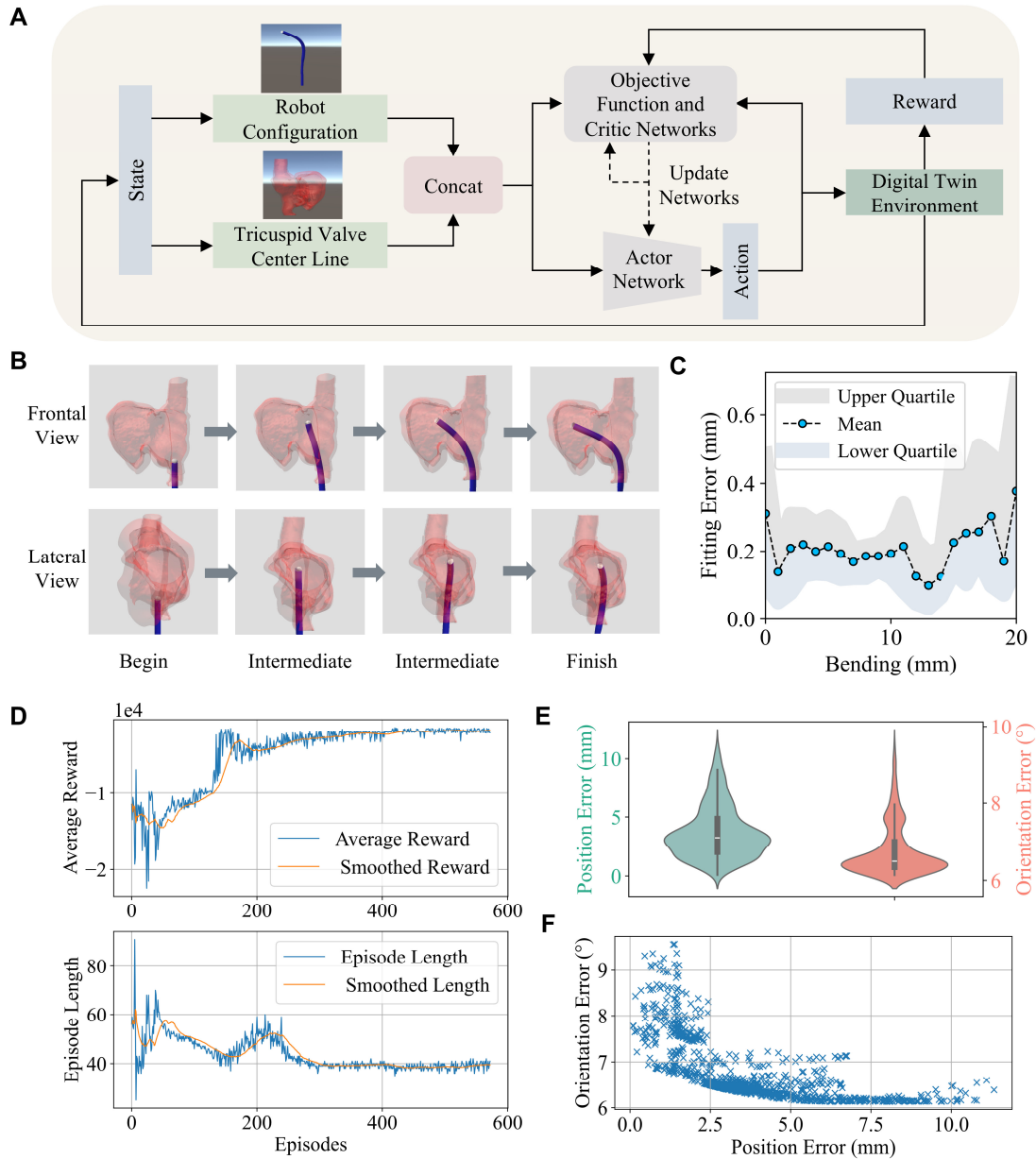

**Fig. 3. Autonomous intracardiac localization strategy based on reinforcement learning.** (**A**) Reinforcement learning-based framework for path planning. (**B**) Autonomous localization process performed in digital twin environment from two views, frontal and lateral. (**C**) Error distribution statistics for neural network fitting of delivery catheter kinematics. (**D**) Changes in average reward and episode length during training of reinforcement learning algorithm. (**E**) Position error and orientation error obtained statistically by the autonomous localization algorithm. (**F**) Statistics on the relationship between positional and orientational error of autonomous localization algorithm.

**Autonomous policies for intracardiac localization**

The most complex and time-consuming part of the TTVR procedure is the localization process, where the surgeon has to manipulate the delivery catheter to accurately align the tricuspid valve, a step that is crucial to the success of the subsequent surgery. Manual manipulation is highly dependent on the surgeon's skill and experience, as the surgeon needs to continuously adjust the three degrees of freedom of overall translation, bending, and overall rotation to ensure that the end of the instrument passes through the annulus accurately. To address these clinical challenges, this study proposes a path planning method based on digital twin and reinforcement learning. The algorithmic framework is shown in Fig. 3(A). We adopt the Soft Actor-Critic (SAC) algorithm (54, 55), a stable off-policy approach to guide the robot to complete the localization process with the most optimal trajectory possible while avoiding collision with the inner wall of the patient's heart.

To validate the effectiveness of the algorithm, we import pre-operative data of a real patient into our pre-planning platform, which contains accurate modelling of the anatomical structure and surgical instrument. Fig. 3(B) shows the result of the autonomous policy for localization in the digital twin environment, observed from both the frontal and lateral views of the heart. The localization process starts after the delivery catheter reaches the superior vena cava, and the instrument would adjust the bending and rotation degrees of freedom in time during delivery so that it can smoothly cross the tricuspid valve from the right atrium into the right ventricle (movie S3).

Since the delivery catheter contains a multilayered structure of outer sheath, inner core tube, and bending tube, the modelling of the bending degrees of freedom appears to be more complicated. For this reason, we discretize the catheter shape into 100 points by optical tracking and use a multilayer neural network to fit the coordinates of each point of the instrumented catheter shape at different values of the bending degrees of freedom. Fig. 3(C) demonstrates the fitting error of the neural network for predicting each point on the catheter for a specific value of the bending degrees of freedom. The average fitting error is 0.24 mm, which is 2.1% of the diameter of the instrumented catheter (11.4 mm). Fig. 3(D) shows the variation of average reward and episode length during the training of the algorithm. After about 400 rounds, the algorithm gradually converges and both average reward and episode length remain stable and no longer change over time. After the algorithm converged, the localization algorithm is executed 1000 times from a random initial position in the digital twin environment, and the statistically obtained position and orientation errors for catheter's tip localization are $3.7 \pm 2.26$ mm and $6.78 \pm 0.68°$, respectively, as shown in Fig. 3(E). Given the under-actuated nature of the localization process, it is difficult to achieve low positional and orientational errors at the same time. Fig. 3(F) statistically shows the distribution of the two types of errors, and it is found that the maximum value of the positional error is about 10 mm, and the maximum value of the directional error is about 10°, which are both within the acceptable range.

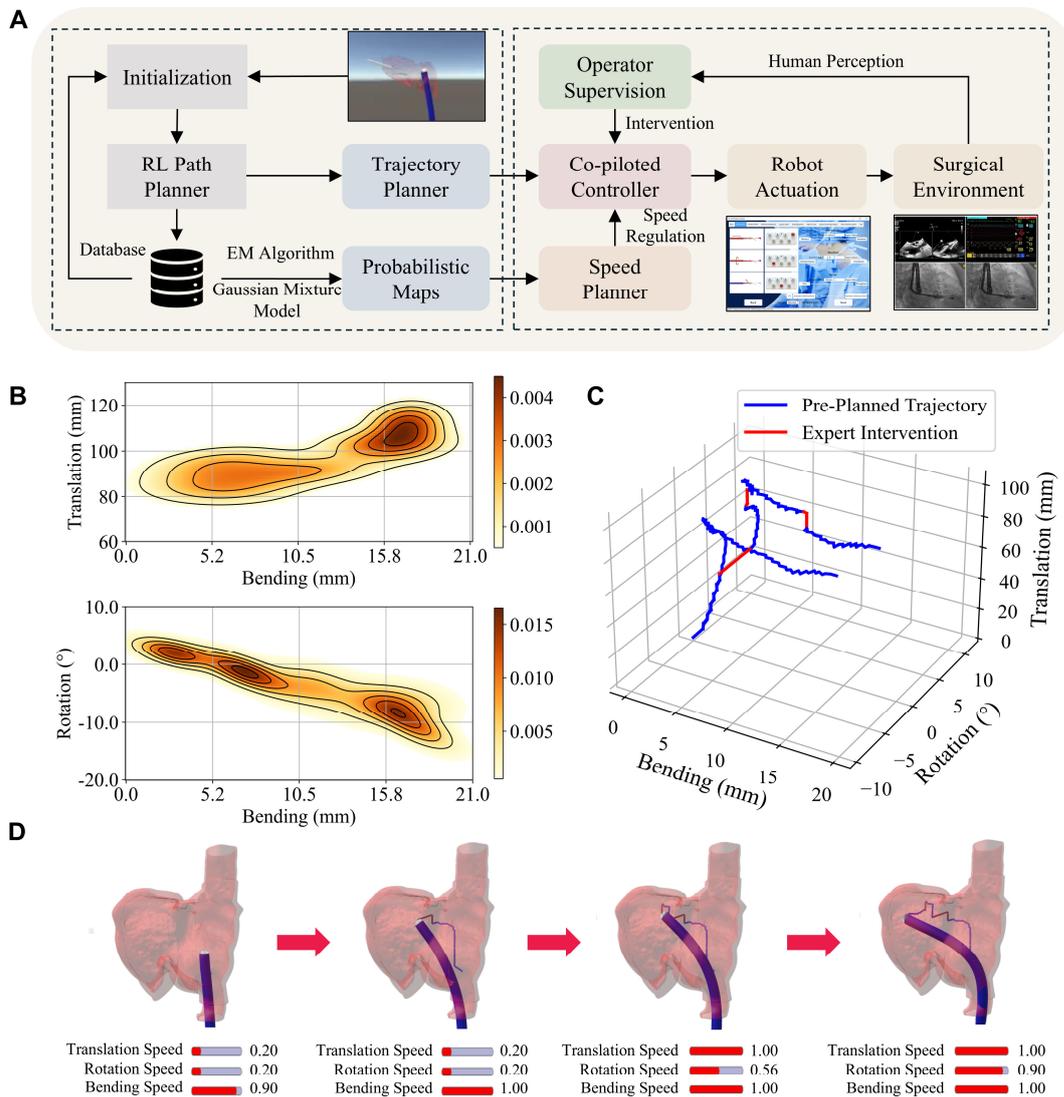

**Fig. 4. Demonstration of probabilistic maps and co-piloted navigation.** (**A**) Overview of the human-robot co-piloted navigation control strategy, which randomly initializes a large number of different delivery catheter positions in a virtual environment and performs reinforcement learning (RL)-based path planning to obtain the probabilistic maps. During the procedure, the controller operates based on a trajectory from the ideal initial position, while the operator supervises the whole process and can intervene at any time to make adjustments to the trajectory, and the controller limits the operator's interventions based on the probabilistic maps. (**B**) Translation-Bending and Rotation-Bending probability maps. (**C**) The pre-planned trajectory used and the actual trajectory of the delivery catheter under operator's intervention obtained from a simulation experiment. (**D**) The relationship between the position of the heart and the delivery catheter after three interventions by the operator and the final localization result, with the speed scales automatically regulated by the probability maps for the three degrees of freedom.

Page **9** of 36

**Probabilistic maps and co-piloted navigation**

In the real surgical scenario, after the delivery catheter enters the heart from the superior vena cava, there is a large uncertainty in its initial location, which limits us from directly applying pre-planned path. To address this problem, we propose a co-piloted control strategy as shown in Fig. 4(A). In the digital twin environment, we use reinforcement learning for motion planning against a large number of randomized heart-catheter initial locations. Subsequently, we fit the translation-bending and rotation-bending distributions separately using Gaussian mixture models to obtain the probability values of the occurrence of each joint parameter in the joint space, and we define these distributions as probabilistic maps, as shown in Fig. 4(B).

When operating the robot, the co-piloted control strategy uses a pre-planned trajectory based on a desired initial position allowing the operator to manipulate three degrees of freedom simultaneously. The operator can intervene on a single degree of freedom at any time when the trajectory needs to be adjusted, and the corresponding interventions are constrained by the probability map. When the operator manipulates the robot joints to move in the direction with a smaller probability value in the probability map, the robot's movement speed is slowed down; conversely, the normal movement speed is maintained. After the expert intervention, the trajectory is updated according to the intervention. We conducted experiments in the digital twin environment based on a real patient data to demonstrate the implementation of the co-piloted approach. Fig. 4(C) shows the pre-planned trajectory used by the controller and the actual trajectory of the catheter movement with the intervention of an operator. Fig. 4(D) shows the positional relationship between the heart and the delivery catheter and the final localization result after three interventions during the experiment. The velocity constraints, expressed by scale, of each step regulated by the probability map were also recorded.

It can be seen that after the delivery catheter entered the heart and moves for a period of time according to the pre-planned trajectory, the operator intervened to operate the rotational degree of freedom to increase, and when the operator increased the angle by a large amount, the speed limit of the rotational degree of freedom decreased to 0.20 times the normal speed, while the speed scale of the bending was always 1.00, which suggests that the operator should handle the bending degree of freedom at this time or continue to execute the pre-planned trajectory. In the subsequent localization operation, the expert intervened to modify the translation and bending degrees of freedom respectively. The above demonstration intuitively illustrates that co-piloted navigation allows efficient intervention in high probability regions and slow intervention in low probability regions, thus achieving a balance between autonomy, effectiveness and safety.

**Validation of the effectiveness in phantom**

To validate the effectiveness of the robotic delivery and valve release system, we invited an experienced clinician to perform a complete tricuspid valve implantation procedure in a self-designed heart phantom using the most basic master-slave control method. As shown in Fig. 5(A), the experimental scenario consists of a valve intervention simulation phantom constructed based on real patient anatomy, with a built-in flexible material to simulate the tricuspid valve and a composite material to assist septal fixation. Meanwhile, three cameras were deployed inside and outside to simulate the intra-operative DSA and TEE as the operator's guidance, which is shown in Fig. 5(B). Fig. 5(C) shows the state of the robot end at different phases of the surgery. As can be seen, the valve was successfully released and fixed to the septum by the anchorage and the leaflet fixation pin was successfully attached to the leaflet. We further analyzed basic data on robot manipulation, of which Fig. 5(D) illustrates the statistics on the percentage of the use of each degree of freedom throughout the surgical simulation process, and Fig. 5(E) illustrates the measured range of motion of each degree of freedom versus the maximum allowed range of motion in the full surgical simulation process. A video recording of an example in-vitro phantom experiment is shown in movie S4.

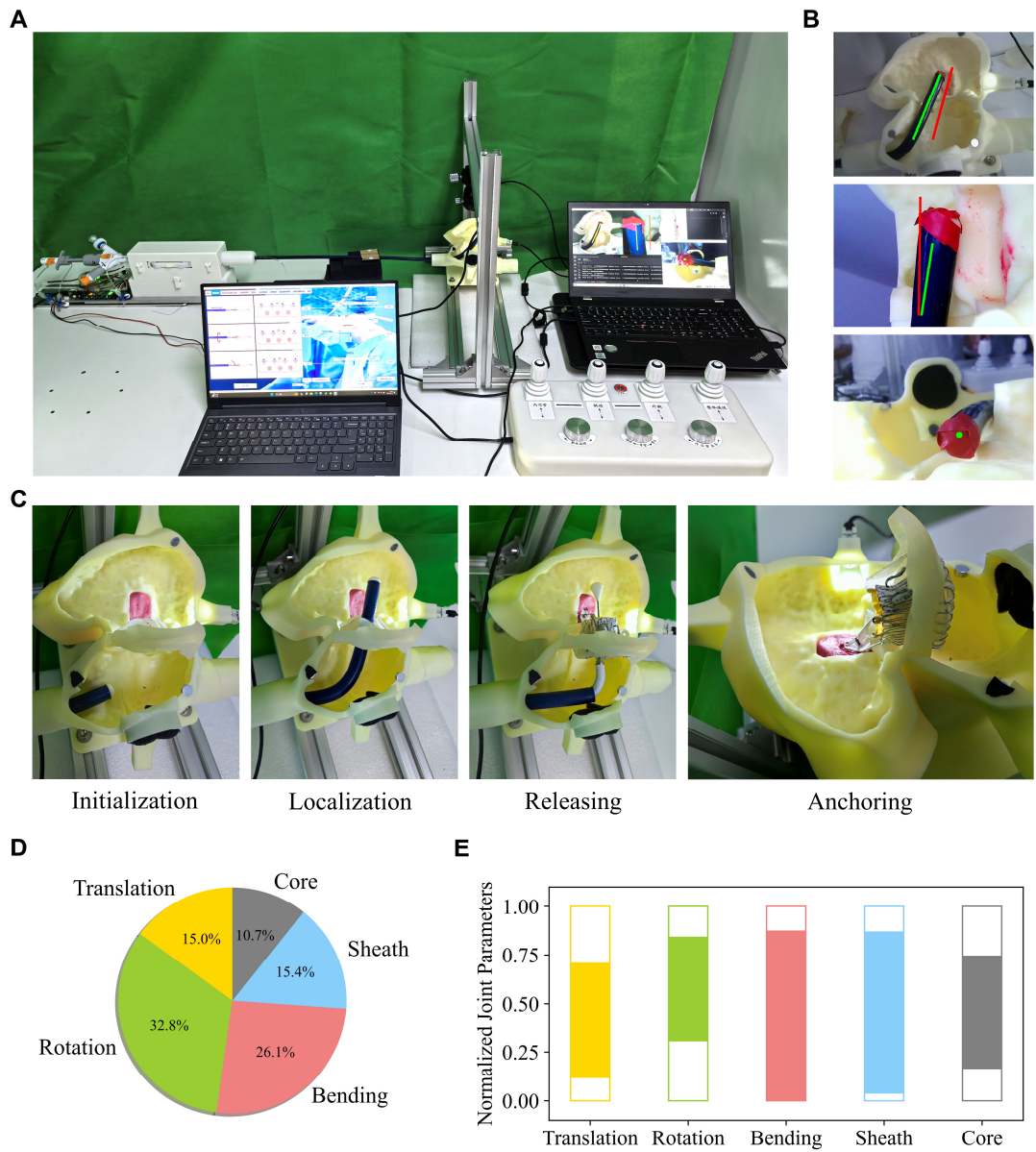

**Fig. 5. Phantom experimental scenario and results of basic tests of the robot's effectiveness.** (**A**) The experimental setup for the phantom tests. (**B**) The camera view used to guide operator feedback in the phantom experiment. From top to bottom are top view, sagittal view and axial view, where the red line in the first two is the desired target position and the green line is the current position. (**C**) The state of the robot end at different phases of the surgery, where the overall bending caused by the bending tube to achieve localization and the valve release caused by the withdrawal of the outer sheath and inner tube can be observed, respectively. (**D**) Statistics on the percentage of the use of each degree of freedom

throughout the surgical simulation process. (**E**) Measured range of motion of each degree of freedom versus the maximum allowed range of motion in the full surgical simulation process, where each degree of freedom is normalized by the maximum allowed range of motion.

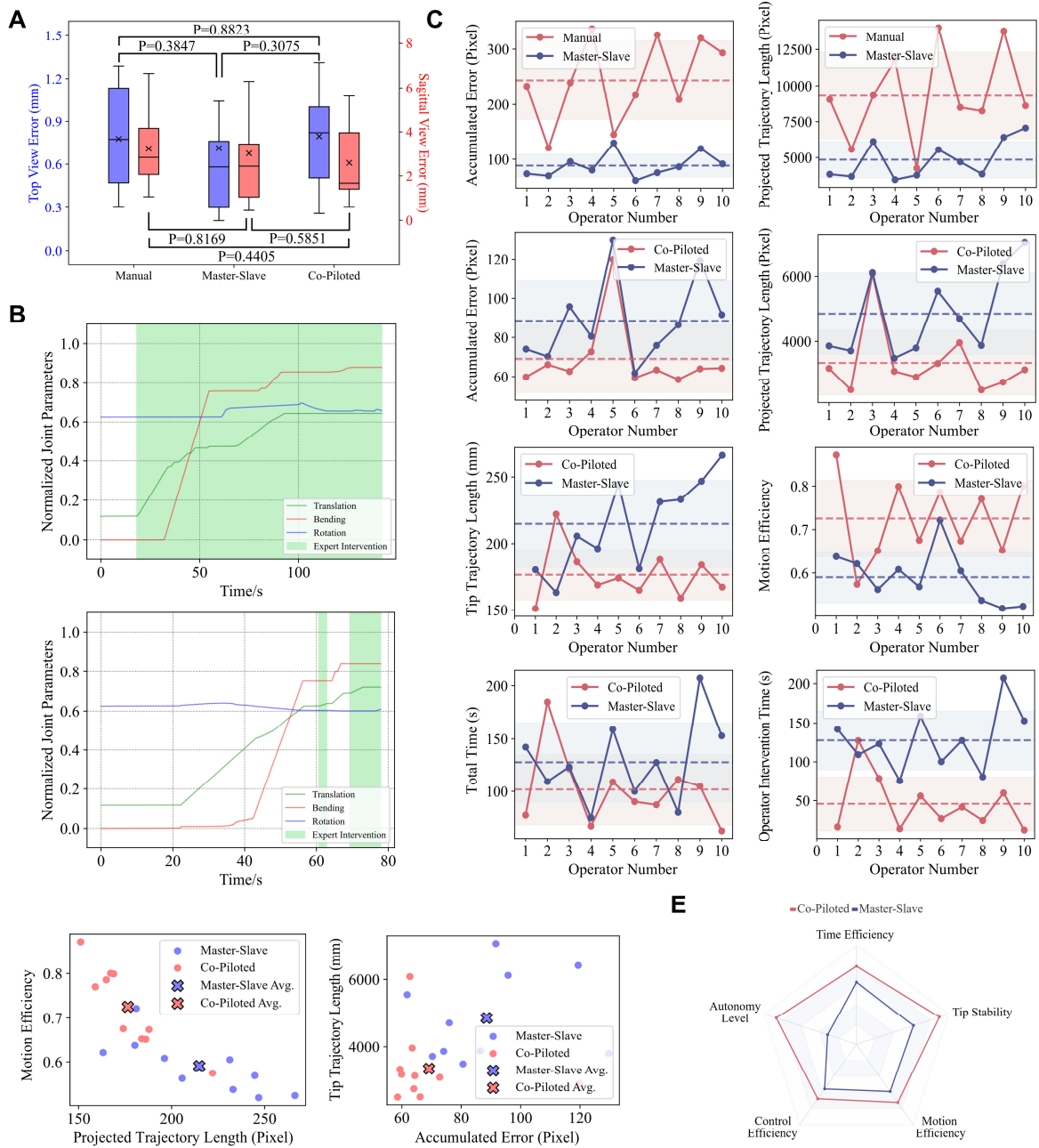

**Fig. 6. Performance results of manual operation, master-slave operation and co-piloted control in phantom experiments.** (**A**) Accuracy statistics of the three manipulation methods in the localization phase. (**B**) An operator case of the joint trajectory that completes the localization process under master-slave and co-piloted control, where each degree of freedom is normalized by the maximum allowed range of motion. (**C**) Comparison of evaluation parameters under three control modes for ten operators. (**D**) Coupling analysis of two sets of parameters reflecting the trajectory of the delivery catheter. (**E**) Evaluation of the robot's capabilities for the two control methods based on the above performance.

**Comparative study of different manipulation methods**

The goal of the second experiment conducted in phantom is divided into two main points, one is to consider the direct manual operation of the delivery catheter as a baseline to compare the performance of indirect operation through the robot, and based on this, the second is to compare the basic master-slave control of the robot with our proposed co-piloted control to explore the role of autonomy assistance. We focus on the localization phase in our comparison experiment, as it has the longest stroke and the most complex movement in the whole surgical process. In the experiment, the performance of ten short-trained novices operating under the three controls was recorded. As with real surgery, there is uncertainty about the initial location in each instance. Firstly, we compared the effectiveness of the three control methods, using the tracking of the delivery catheter by two cameras as the evaluation to calculate the error between the location of the delivery catheter and the ideal target after the completion of localization phase. The ideal location was calibrated by the experienced surgeon manually in advance as the true value. Here, operational data under different control strategies for all operators are shown in fig. S1.

Figure 6(A) shows the error statistics, and the hypothesis test demonstrates that there is no significant difference between the three (The p-value obtained from the two-by-two analysis is shown in Fig. 6(A)), indicating that the effect of direct control by human hands can be achieved under the indirect control of the robot. For the two control modes of the robot, Fig. 6(B) shows the joint trajectories, and it can be seen that the manual intervention is significantly reduced under the co-piloted control. Fig. 6(C) analyses the control methodology by means of a series of quantitative parameters, which are defined in detail in the materials and methods section. Comparing the manual and robotic master-slave control, it can be seen that the robot outperforms the human hand in the parameters of accumulated error and projected trajectory length analyzed by the camera tracking ($p = 0.0001$ and $p = 0.0008$), which suggests that the localization phase can be more efficiently and stably accomplished under robotic control. In addition, we also note a more uniform performance across operators under robotic manipulation, with the accumulated error and projected trajectory length being $244 \pm 71$ pixel (mean ± std) and $9311 \pm 2993$ pixel (mean ± std) under manual manipulation and $89 \pm 20$ pixel (mean ± std) and $4867 \pm 1251$ pixel (mean ± std) under robotic manipulation, respectively.

Comparing the master-slave and our proposed co-piloted control, we can see that the co-piloted control performs better in the parameters of accumulated error and projected trajectory length obtained from the camera tracking analysis ($p = 0.0113$ and $p = 0.0046$), and co-piloted control also performs better in tip trajectory length and motion efficiency parameters obtained through robot joint parameter recording and kinematic calculation ($p = 0.0214$ and $p = 0.0019$). Here, tip trajectory length is a reflection of control efficiency at the operator's control end, and motion efficiency is a reflection of the delivery catheter tip reaching its target. The above performance is also reflected in the time parameters, where it can be seen that the co-piloted control, although not significantly different from master-slave in terms of total time ($p = 0.1671$), is significantly smaller than the latter in terms of operator intervention time ($p = 0.0006$). Here, the operator intervention time is also an indicator of the autonomy level. In addition, Fig. 6(D) shows a coupled presentation of the two parameters obtained from the analysis under camera tracking and the two parameters obtained from the analysis of the robot's joint parameters and kinematic calculations, which further demonstrates the advantages of co-piloted control. Based on the above findings, we have made a visual representation of the capabilities of master-slave and co-piloted control through radargrams, which are displayed in Fig. 6(E).

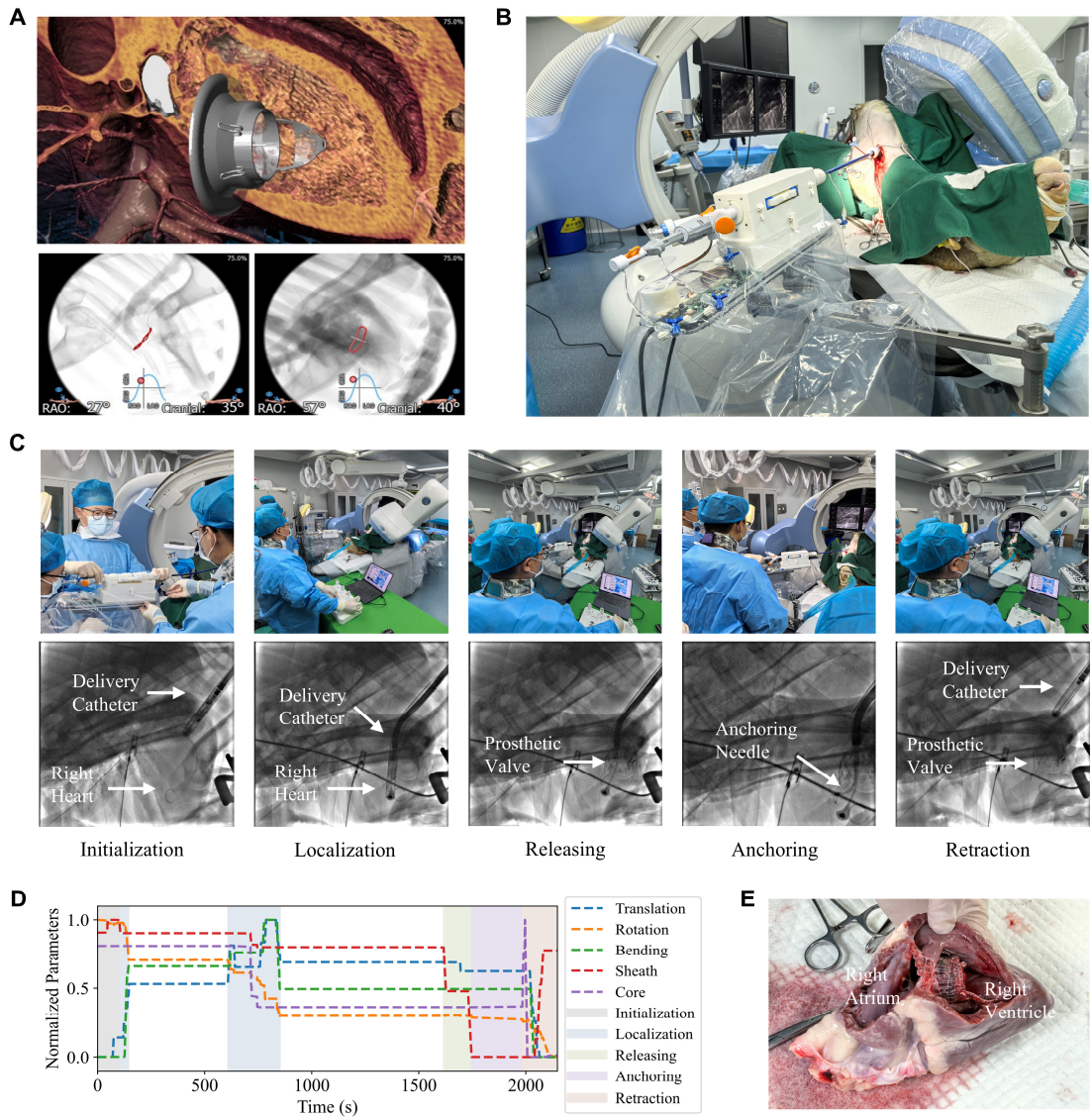

Fig. 7. **Demonstration of the effectiveness of robot-assisted tricuspid valve replacement in a full-flow in vivo trial.** (**A**) Pre-operative evaluation based on 3mensio software and CT data, where the top image shows a rendering of the virtual implantation of the prosthetic valve and the bottom images show the recommended range of projection angles for DSA imaging. (**B**) The animal test scene contains the passive stabilizer, the robotic drive isolated by a sterile cover, the sterilizable instrument box and the delivery catheter with the established transjugular access. (**C**) Records of the entire surgical process in the phases of initialization, localization, releasing, anchoring, and retraction. The upper figure shows the configuration of the master-slave control of the corresponding stage, and the lower figure shows the DSA images of different stages, which can be seen in the entry and position adjustment of the delivery catheter, the release of the valve and its fixation through the anchoring needle, and the final withdrawal. (**D**) Variation of robotic joint parameters throughout the surgical procedure, where each degree of freedom is normalized by the maximum range of operator's motion. (**E**) Dissection of the right heart at the end of the study demonstrates the successful placement of the prosthetic valve.

**In-vivo validation of full surgical process**

To validate the valve implantation and human-robot interaction under real cardiac conditions, we performed animal experiments. We chose sheep for our animal experiments because it has a slender neck, which makes it easier for the catheter to enter the right atrium through the superior vena cava. Our first experiment focused on verifying the effectiveness of the full robotic process, thus the standard master-slave control was utilized. The surgical procedure began with a pre-operative evaluation of the heart to determine the size of the prosthetic valve. Computed Tomography (CT) images were used to evaluate key parameters such as valve annulus dimensions, the angle of the septum with respect to the bed plane, and the required angle of the delivery catheter with respect to the bed plane, using the commercially available software 3mensio (Pie Medical Imaging, Utrecht, The Netherlands), the results of which are shown in Fig. 7(A). In the experiment, the prosthetic valve was attached to the end of the delivery catheter prior to surgery (fig. S2).

During the procedure, the delivery catheter was manually inserted into the jugular vein to reach the entrance of the right atrium and then connected to the robot. Thereafter, the robot takes over the subsequent control, and the configured scene is shown in Fig. 7(B). As shown in Fig. 7(C), the first phase was initialization, in which the robot's delivery and overall rotation degrees of freedom were adjusted to bring the end of the instrument to the initial position at the cardiac inlet. In the phase of localization, the end of the delivery system was aligned with the vertical axis of the valve plane by adjusting the translation, bending, and overall rotation. In the phase of releasing, the prosthetic valve was first exposed by withdrawing the outer sheath, and then the outer sheath and inner core tubes were adjusted together to allow the valve to move back until the valve annulus were hooked onto the leaflet fixation pins. Finally, the outer sheath was withdrawn further to fully release the prosthetic valve. In the phase of anchoring, the needle was manually pushed out through the delivery catheter and the prosthetic valve was fixed. Finally, the delivery catheter was retracted and removed from the body.

Figure 7(D) shows the variation of the robot joint parameters throughout the procedure by the experienced operator, where each degree of freedom is normalized to the maximum range of operator's motion. We can see that the degrees of freedom used by the operator at each stage of the master-slave operation are in accordance with the design concept of the robot and delivery catheter. The main surgical time for the whole procedure under master-slave operation was 35 minutes. Fig. 7(E) shows the dissection results of the heart at the end of the experiment, and the experimental results show that the robot successfully implanted the prosthetic valve in the correct position. The whole experimental results validate the safety and effectiveness of the proposed robot and the standard master-slave control. The video recording of the whole procedure can be found in movie S5.

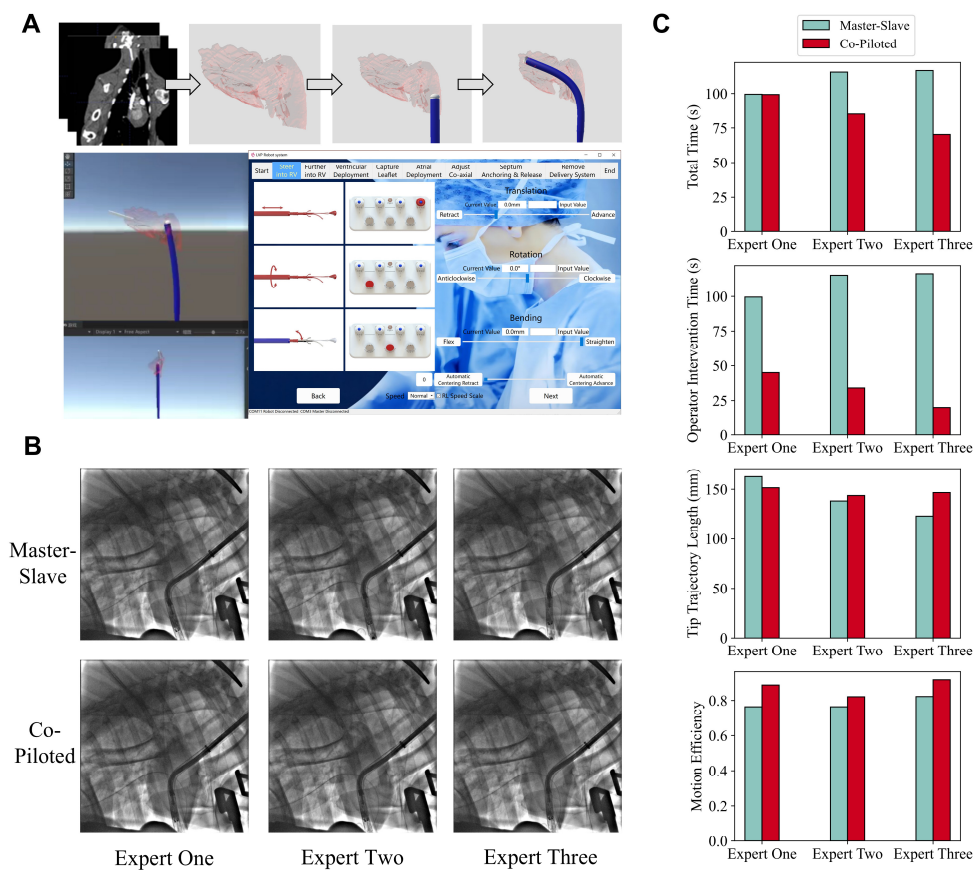

**Fig. 8. Validation of the effectiveness of hybrid enhanced intelligence formed by probabilistic maps and co-piloted control during in vivo testing.** (**A**) Demonstration of pre-operative planning and autonomous learning, in which the upper figure shows the CT scan of the sheep, segmentation of the critical region of the right heart, initialization of reinforcement learning and the final result, respectively, and the lower figure demonstrates the coupling of the robot manipulation software with the digital twin platform, in which the operator can virtually validate and confirm the co-piloted control before the procedure. (**B**) Localization results of the delivery catheter in the DSA imaging obtained by three different experienced operators under master-slave and co-piloted control. (**C**) Typical parameter performance of three different experienced operators under master-slave and co-piloted control, which includes the overall length of time to complete the localization phase, the length of the operator's intervention time at the master console, the tip trajectory length and the motion efficiency calculated based on the joint parameters.

**In-vivo demonstration of co-piloted localization**

On the basis of the above experiment, our second sheep animal trial focused on experimenting with hybrid enhanced intelligence-driven human-robot co-piloted methods in the challenging localization phase of surgery. In addition to the pre-operative valve implantation analyzed by 3mensio software, we manually segmented the superior vena cava, right atrium, and right ventricle of the sheep by ITK-SNAP (version 4.0.2) and reconstructed their anatomical structures in three dimensions. Fig. 8(A) illustrates the above analysis process, where the top figure shows the CT scan of a sheep, the segmentation of the critical region of the right heart, the initialization of reinforcement learning and the final result, respectively, and the bottom figure shows the coupling of the robot manipulation software with the pre-planning platform, where the operator can virtually validate the co-piloted control.

The surgical preparation process for the second sheep was the same as the first one, with the difference that we only tested the localization step in the second sheep in the following process, which is the difficult part of the surgery and the main problem that the hybrid enhanced intelligence-driven co-control is trying to solve. After the delivery catheter was placed in the cardiac chambers and the system was switched to robotic control, we asked the operator to perform the localization operation using standard master-slave control and co-piloted control, respectively. Once the operation was completed, the delivery catheter was deliberately withdrawn, and the next operator was invited to perform the same operation. Considering the length of the procedure, animal tolerance, and operator confidence, three experienced operators were selected to complete the experiment.

Figure 8(B) shows the results of the positioning of the delivery catheter in the DSA imaging obtained by three experienced operators under master-slave and co-piloted control. As verified by the images, the positioning of the delivery catheter was successfully accomplished by all three operators under the two different maneuvers. Fig. 8(C) shows the analysis of the robot joint trajectories for both control methods. It can be seen that in terms of the total duration of the localization process, all three operators completed it faster or in the same amount of time by the co-piloted approach. Considering the length of the operator's intervention time at the master console, the co-piloted maneuver significantly reduces the manual actions required by the operator. In terms of tip trajectory length, the difference between the two controls was not significant, and in terms of motion efficiency, all three operators performed slightly better under co-piloted control.

## DISCUSSION

In this study, we propose a novel design of a tricuspid valve replacement robot and a co-piloted control strategy driven by hybrid enhanced intelligence. Based on this, a complete clinical solution is proposed, and the feasibility and effectiveness are fully verified by phantom experiments and animal tests. To the best of our knowledge, this is the first robotic system developed for tricuspid valve replacement that successfully integrates intelligent assistance strategies and completes the whole process of in-vivo testing. Therefore, this study is a significant landmark for the development of robot-assisted cardiac interventions. As TTVR is still a procedure that has not yet been widely disseminated, our robotic and intelligent assisted solution is expected to be introduced into the clinic in parallel with the manually operated instruments and operating guidelines. This is also of great value for the development of new procedures through medical robotics.

Our proposed robotic system takes into account the complexity of valve release and fixation, the need for instrument sterilization, and the transition and consistency between manual and robotic manipulation, making it a design that fully meets the clinical needs. Benchtop experiments have shown that our design provides a reasonable workspace to meet the requirements of the surgical procedure, whilst providing sufficient stiffness in the event of perturbation being applied. Phantom and animal studies have also shown that the valve can be successfully implanted with this workspace and stiffness design.

On the drive side of the robot, we have been able to separate and integrate the instrument, instrument case and drive unit into a very small space. This solves the sterilization problem and allows the operator to switch between manual and robotic control. The clinical utility of these maneuvers was well established in animal trials. In our design, the configuration of the main operator side of the robot is highly consistent with the logic of manually operating the delivery catheter. One of the experiments conducted in the phantom showed that the critical performance of the operator-controlled robot is comparable to that of a manually operated delivery catheter, and that the robot provides better stability by ensuring that the valve does not lose its position when aligned with the valve annulus, and by allowing the operator to move away from radiation, which is not possible with manually operation. Both are of great concern to clinicians.

Further, we explore the use of hybrid enhanced intelligence to improve robot autonomy and fully validate that this is a clinically acceptable and effective strategy. There are three main dimensions of this intelligence, the first being the control system combined with software-guided structured surgery, where the operator is prompted to operate at the current step and is only allowed to move as many degrees of freedom as are appropriate for that step. The second is the spatial understanding of the surgery space brought by autonomous planning. We make full use of the pre-operative information to build an efficient digital twin-based pre-planning platform and generate individual probabilistic maps using reinforcement learning and Monte Carlo approaches. Finally, the operator-centered co-piloted control, where the operator efficiently controls multiple degrees of freedom in conjunction with the automatic-generated trajectory and is still allowed to stop, continue and intervene in a single degree of freedom and at any point, while the probabilistic map regulates the response speed in the background. The results of phantom and animal experiments show that the method can autonomously and efficiently complete the planning of localization, and operators were able to achieve better performance under co-piloted control. This strategy enhances autonomy while preserving the central role of the operator and thus clarifying responsibility and safety.

There are still several aspects of this study that need further improvement. At the design level, the delivery catheter currently only has the ability to bend in one direction, and the introduction of bending in the other direction would further improve the precision of the localization step but would also imply a more complex multi-layered concentric tube structure, the necessity for which will need to be further justified by future clinical evidence. On the control side, the introduction of DSA and TEE feedback into hybrid intelligence will further improve the autonomy of the robot, which depends on obtaining more image data for training and further clarifying the judgement logic through clinical research, which will also be the focus of our next work. In addition, we will continue to expand the number of animal experiments and the number of operators to further validate and optimize the co-piloted control strategy.

## METHODS
### Study Design and Statistical Information
The objective of this study was to develop a robotic surgical system with a human-robot co-piloted control strategy for transcatheter tricuspid valve replacement and validate its clinical effectiveness towards autonomy. The experiments were divided into two parts, including vitro experiments and in-vivo animal testing which conducted in accordance with the protocol approved by the Laboratory Animal Management and Use Committee of the host animal testing centre (Gateway, Beijing, China, Approval Number: BJ2024-08001 for the experiment case PAE0526).

The purposes of vitro experiments were to demonstrate the effectiveness and safety of the proposed robotic surgical system and validate different manipulation methods, including manual operation, master-slave control and co-piloted control, to explore the role of autonomy assistance. First, an experienced clinician performed a complete tricuspid valve implantation procedure in a self-designed heart phantom to validate the effectiveness of the robotic delivery and valve release system. Subsequently, we focused on the localization phase in our comparison experiment, as it has the longest stroke and the most complex movement in the whole surgical process. The short-trained participants (n=10) were required to finish this task via three control methods mentioned above.

In the in-vivo animal testing, we further validated the clinical utility of the proposed robotic surgical system under real cardiac conditions. Sheep were selected for the experiments because it has a slender neck, which makes it easier for the catheter to enter the right atrium through the superior vena cava. One experienced clinician was required to finish the full surgery process on the first one, while three experienced clinicians were participated in the testing on the second sheep animal trial, focusing on experimenting with hybrid enhanced intelligence-driven human-robot co-piloted methods in the challenging localization phase of surgery, with sheep tolerance considered.

**Implementation of the robotic system**
The robot uses four brushless DC motors with built-in planetary gearboxes (MG4010E-i10v3, Ling Kong Technology, Suzhou, China) to drive the catheter handwheels, one direct-drive brushless DC motor (MF4005v2, Ling Kong Technology, Suzhou, China) to drive the overall catheter translation, and one miniature brushless DC motor with a gearbox (CHF-GM12-N20VA ABHL, Chi Hai Motor, Shenzhen, China) to drive the gripping jaw wheel. The robot's base plate and bearing parts are made of metal, and the housing material is made by 3D printing with light-curing materials. The size of the drive part is 400*100*143 mm, and the length of the delivery catheter is 1,100 mm. The master control console consists of four self-recovery joysticks (SMC35B, Shenzhen Xiao Long Electrical Co. Ltd, Shenzhen, China) and angle sensors (P3022-V1-CW360, PandAuto Co. Ltd, Shanghai, China) combined with three self-recovery knobs made from springs and the main structures are made using 3D printing with light-curing materials. The robot has a custom-made controller board, which is used for motion control, communication with the host computer, and safety monitoring of the motion status. The Qt-based interface includes the operation steps of each surgical phase and restricts the unnecessary operations in each process. The control architecture for different levels of the proposed robotic system is shown in fig. S3.

**Digital twin and pre-planning platform**
The pre-planning platform not only provides an ideal environment for the training of reinforcement learning algorithm, but also enables the surgeons to familiarize themselves with the surgical process before the surgery and virtually attempt different human-robot co-piloted control methods. We uniformly affixed eight markers in the catheter's bendable region and tracked them using an optical tracking system (OptiTrack V120-Trio, NaturalPoint Inc., Oregon, USA), so that we can obtain and reconstruct the catheter's shape using a lightweight neural network. Information on the above process can be found in fig. S4. Then we utilize the Unity (Version 2022.3.29) software to construct the pre-planning platform, incorporating the visualization of patient's heart based on preoperative CT and reconstructed catheter. Once the clinician finishes calibrating the centerline of the tricuspid valve, the engine is able to compute and display the precise positional relationship between the catheter and the heart, which is used to output the State and Reward information required for reinforcement learning algorithm, providing important support for decision-making. Information on the construction of the pre-planning platform described above can be found in fig. S5.

**Reinforcement learning based path planning**
Reinforcement learning learns optimal decision-making strategies through interaction with the environment (*56*). In order to adapt to the complex anatomical environment faced in path planning, we use a Soft Actor-Critic (SAC)-based reinforcement learning algorithm for automatic localization policy generation.

Two parts are considered as the state in the algorithm: firstly, the current joint parameters of the robot and secondly, the coordinate information of the centerline of the tricuspid valve. Actions are defined as the amount of adjustment of the robot's three joint parameters during localization. Each episode the robot iteratively performs the localization steps until the end of the instrument reaches a specified position or the time step exceeds a specific threshold or the catheter is at maximum bend. Inspired by (*55*), the design of the reward function in the algorithm is structured to optimize catheter navigation and operation efficiency, considering several key factors. Firstly, a negative reward, $r_{step} = -50$, is given at each time step to fasten the algorithm convergence. Secondly, to prevent potential injuries, the algorithm introduces penalties, $r_{obstacle} = -300$, when the delivery catheter comes into contact with the heart. Meanwhile, $r_{error}$ is calculated as the negative sum of the Euclidean distances in the X and Y directions from two points ($P^1, P^2$) on the tricuspid valve centerline to the end of the delivery catheter's coordinate system to enhance accuracy. Lastly, upon successful localization of the instrument's end with the tricuspid valve centerline, a positive reward $r_{target} = 300$ is given to reinforce this precise operation. Thus, the reward function is defined as:

$$r_t = \begin{cases} r_{obstacle} + r_{step} & \text{if collision} \\ r_{error} + r_{step} & \text{if timeout or catheter at maximum bend} \\ r_{target} + r_{error} + r_{step} & \text{if catheter reaches the target} \\ r_{step} & \text{otherwise} \end{cases}. \quad (1)$$

**Phantom design and benchtop setup**
Here, the anatomical model was created using 3D printing and included human structures such as the right atrium, tricuspid valve, right ventricle and parts of the superior and inferior vena cava. The tricuspid valve and adjacent muscle tissue of the phantom were designed as separate parts and made from a 50HA (Shore A hardness) white rubber-like material (Agilus30). To replicate the anchoring effect of the prosthetic valve to the myocardium, a composite material containing polymer clay was used in the anchoring area of the right ventricle. To reproduce the image guidance of the real surgery, the phantom was partially made of transparent resin and three cameras (Two internal: Y102, Dan Nan Technology, Shenzhen, China and one external: G200, RMONCAM, Shenzhen, China) were placed inside and outside to provide guidance and record data. The Phantom was further secured to the stand to keep it stable during the experiments. The configuration and details of the heart phantom can be visualized in fig. S8.

**Analysis of user performances in phantom**

We have implemented the tracking of the delivery catheter position to evaluate the robot's motion through a machine vision-based approach using cameras equipped within the phantom. This algorithm consists mainly of threshold segmentation based on the HSV color model and morphological correction based on the erosion and expansion algorithm. Combined with the robot joint motion trajectories recorded by the encoder, our evaluation metrics are defined. The clinician marked the ideal position of the delivery catheter by drawing a straight line in both the top and sagittal camera views. A straight line can be represented by a series of sample points. Top View Error (TVE) and Sagittal View Error (SVE) at time $t$ is defined as:

$$\text{TVE}_t = \frac{1}{N} \sum_{i=1}^{N} \left\| \boldsymbol{p}_{it}^{\text{real}} - \boldsymbol{p}_i^{\text{ideal}} \right\| \tag{2}$$

$$\text{SVE}_t = \frac{1}{N} \sum_{i=1}^{N} \left\| \boldsymbol{q}_{it}^{\text{real}} - \boldsymbol{q}_i^{\text{ideal}} \right\| \tag{3}$$

where $\boldsymbol{p}_i^{\text{ideal}}$ and $\boldsymbol{q}_i^{\text{ideal}}$ represent the coordinates of the $i^{\text{th}}$ sample point on the ideal position line in the top view and sagittal view, respectively. $\boldsymbol{p}_{it}^{\text{real}}$ and $\boldsymbol{q}_{it}^{\text{real}}$ represent the coordinates of the $i^{\text{th}}$ sample point on the real catheter position obtained by the threshold segmentation algorithm at time frame $t$ in the top view and sagittal view, respectively. Accumulated error (AE) is the total error accumulated over all time frames, defined by summing the errors in the top view and sagittal view for each time frame:

$$\text{AE} = \sum_t (\text{TVE}_t + \text{SVE}_t) \tag{4}$$

Projected Trajectory Length (PTL) is a metric used to measure the stability of a robotic end-effector's motion. It quantifies the pixel distance travelled by the end-effector's projected position on the camera sensor during its motion, as the catheter's jitter can cause its end-effector's pixel coordinates to change dramatically within the camera's perspective. PTL is defined as:

$$\text{PTL} = \sum_t \left\| \boldsymbol{Tip}_t^{\text{projected}} - \boldsymbol{Tip}_{t-1}^{\text{projected}} \right\| \tag{5}$$

where $\boldsymbol{Tip}_t^{\text{projected}}$ represents the pixel coordinates of the robotic end-effector at time frame $t$. Tip Trajectory Length (TTL) is a metric that measures the control efficiency of a robot control system by quantifying the path length of the robot's end-effector in three-dimensional space. It is defined as the distance travelled by the robot's end effector as a function of the joint space parameters $c_t$ at time $t$. The definition of TTL is as follows:

$$\boldsymbol{Tip}_t = F(c_t) \tag{6}$$

$$\text{TTL} = \sum_t \left\| \boldsymbol{Tip}_t - \boldsymbol{Tip}_{t-1} \right\| \tag{7}$$

where $F$ is the forward kinematics function of the robot, and it translates the joint space parameters $c_t$ into the corresponding Cartesian coordinates of the end effector. Motion Efficiency (ME) is a key metric for assessing the performance of a robot's motion, focusing on how efficiently the robot completes tasks. ME is defined as:

$$\text{ME} = \frac{\left\| \boldsymbol{Tip}_{\text{final}} - \boldsymbol{Tip}_{\text{start}} \right\|}{\text{TTL}} \tag{8}$$

where $\left\| \boldsymbol{Tip}_{\text{final}} - \boldsymbol{Tip}_{\text{start}} \right\|$ is the straight-line distance between the starting and ending positions of the end-effector, while TTL is the total path length of the end-effector's trajectory. For the performances of the three control methods, our statistical method is as follows: Independent samples T-tests were applied to sets of parameters that showed normal distribution (Shapiro-Wilk test) and equality of variances (Levene test), whereas Mann Whitney tests were used for the remaining sets of parameters that did not show a normal distribution or equality of variances.

**Configuration of in-vivo animal study**

One 78 kg and one 75 kg male sheep were used for this animal experiment. All procedures were conducted in accordance with the protocol approved by the Laboratory Animal Management and Use Committee of the host animal testing centre (Gateway, Beijing, China, Approval Number: BJ2024-08001 for the experiment case PAE0526). Prior to the experiment, pre-operative volumetric data were obtained from CT scans (LightSpeed VCT, GE HealthCare) in both sheep (fig. S10) for pre-operative analysis. During the experiment, each sheep was anaesthetized, and interventional instruments and anatomical structures were monitored with a DSA system (Optima IGS 330, GE HealthCare) and an ultrasound system (Vivid S70N, GE HealthCare) when necessary, throughout the procedure. The prosthetic valve (LuX-Valve, Jenscare Group) was manually assembled pre-operatively and the dimensions were 30-40 mm. After the experiment, the experimental animals were euthanized in accordance with the standard operating procedures. The heart and venous vessels of the experimental animals were dissected, and the valve replacement result was observed and recorded.

**Acknowledgments:** We thank Jenscare and Gateway team for their guidance and support of the in-vitro and in-vivo animal testing in this study.

**Funding:** This research was funded in part by the National Natural Science Foundation of China under grant 62373352, in part by the InnoHK program, and in part by the Jenscare Group.

**Author contributions:**
S.W. contributed to conceptualization, methodology, investigation, visualization, funding acquisition, project administration, supervision, and writing. H.L., Y.X., Z.W., and L.T. contributed to methodology, investigation, visualization, and writing. X.H., X.H.Z., and C.C. contributed to methodology, funding acquisition, supervision, and writing. K.C.Y.S., S.L., F.P., and Z.G.H. contributed to conceptualization and writing. D.C. contributed to investigation and visualization.

**Competing interests:** S.W., Y.X., Z.W., X.H. are listed as inventors of a patent on the mechanical design of the valve intervention robot. The other authors declare that they have no competing financial interests.

**Correspondence** and requests for materials should be addressed to Shuangyi Wang.

**Supplementary Information**
    **Supplementary figures:**
Fig. S1. Operational data under different control strategies for all operators in phantom experiment.
Fig. S2. Preparation and installation of prosthetic valves prior to in-vivo animal experiments.
Fig. S3. Control architecture for different levels of the proposed robotic system.
Fig. S4. Illustration of the detailed approach used to model the kinematics of the delivery catheter.
Fig. S5. Diagram showing the details of the pre-planning platform construction.

    **Supplementary movies:**
Movie S1. Video illustration of the basic principles and main procedure of TTVR.
Movie S2. Animated schematic of the detachable modular design of the proposed robotic drive and consumables.
Movie S3. Demonstration of the autonomous localization strategy obtained through reinforcement learning in the pre-planning environment.
Movie S4. Video recording of an example in-vitro phantom experiment.
Movie S5. Video recording of the in-vivo full-process experimental procedure.

**Supplementary Figures:**

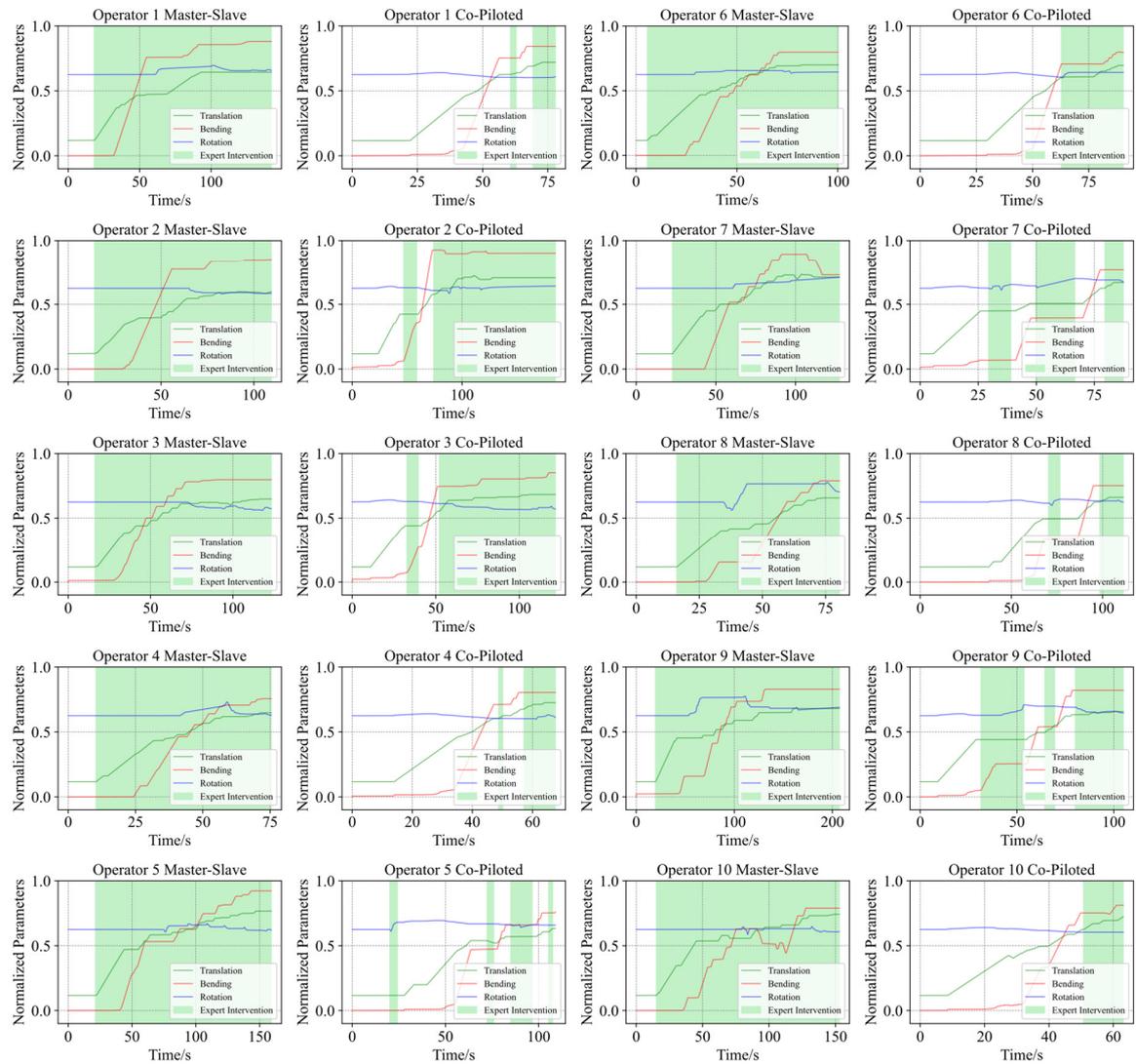

**Fig. S1. Operational data under different control strategies for all operators in phantom experiment.**

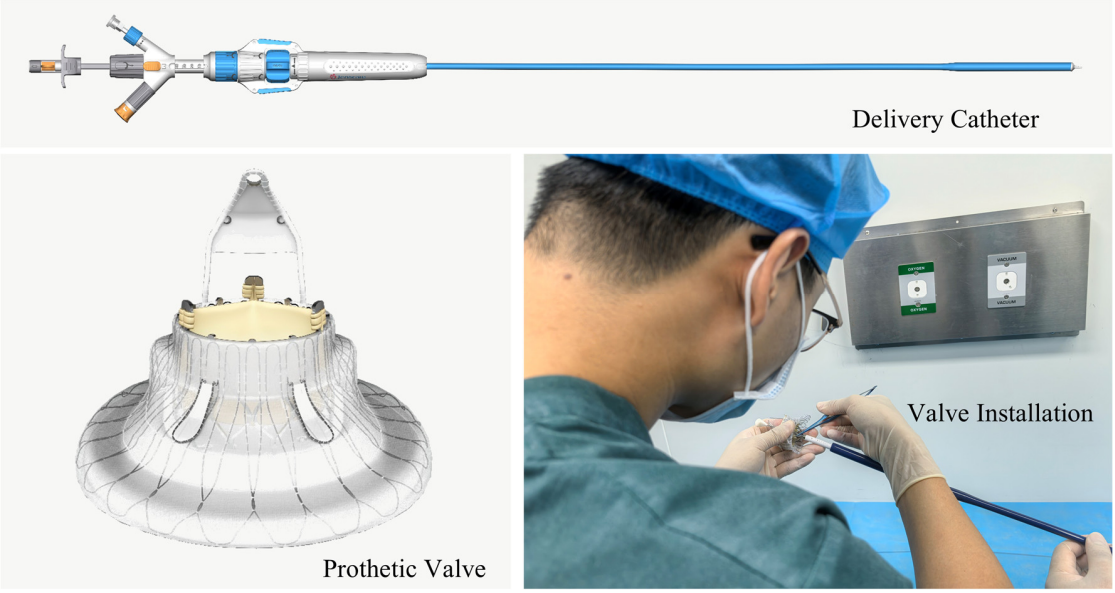

**Fig. S2. Preparation and installation of prosthetic valves prior to in-vivo animal experiments.**

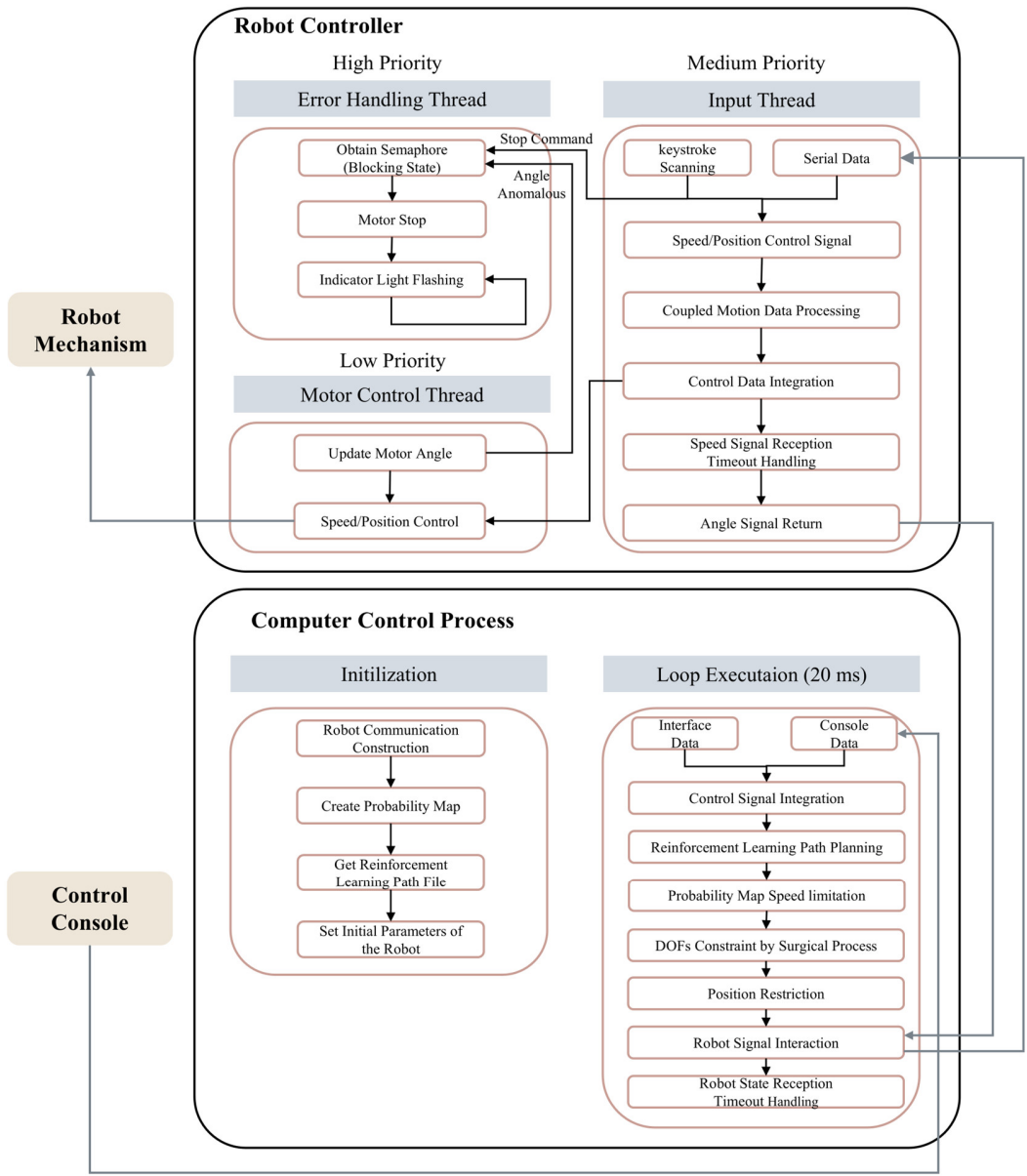

**Fig. S3. Control architecture for different levels of the proposed robotic system.**

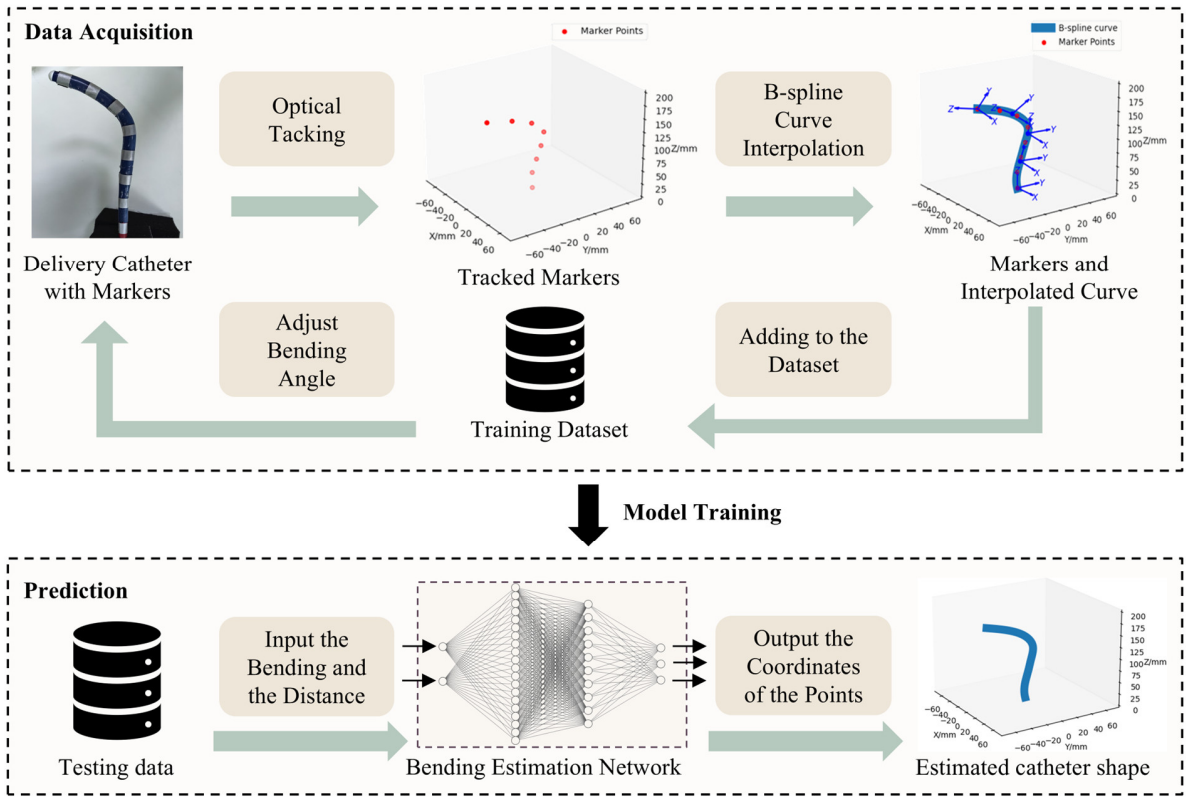

**Fig. S4. Illustration of the detailed approach used to model the kinematics of the delivery catheter.**

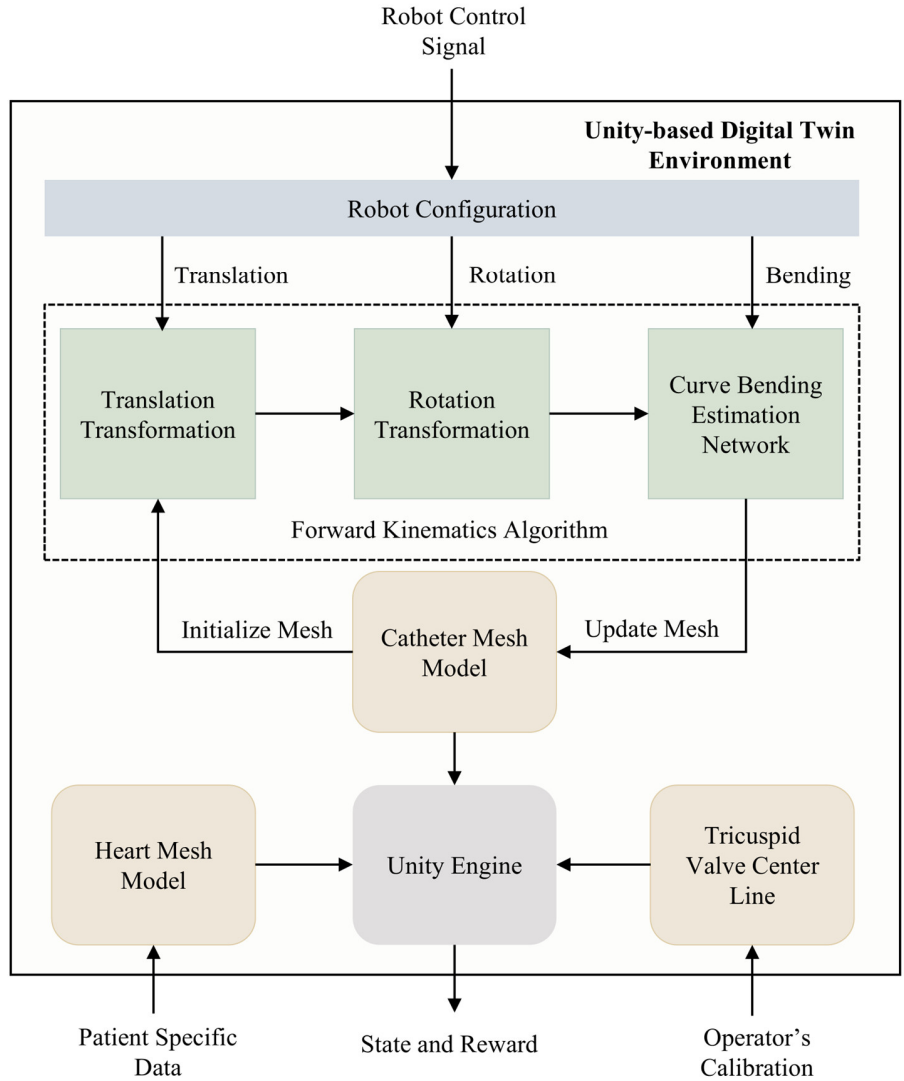

**Fig. S5. Diagram showing the details of the pre-planning platform construction.**

**Supplementary Movies:**

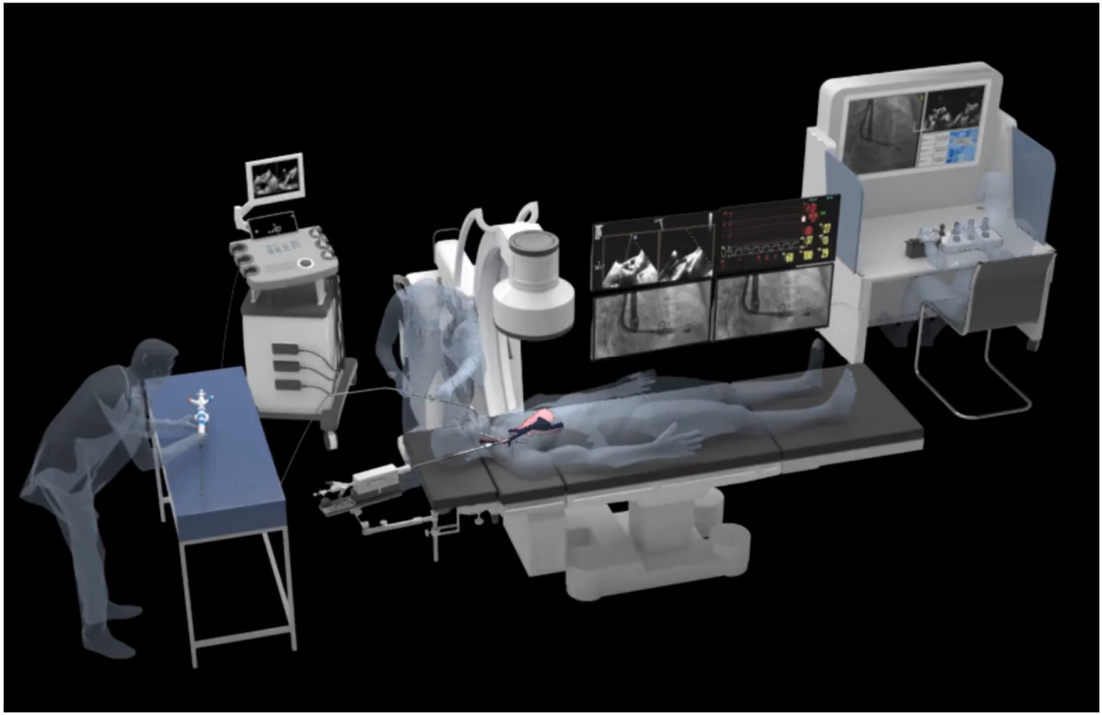

**Movie S1. Video illustration of the basic principles and main procedure of TTVR (See file movie_S1 in the Supplementary Materials).**

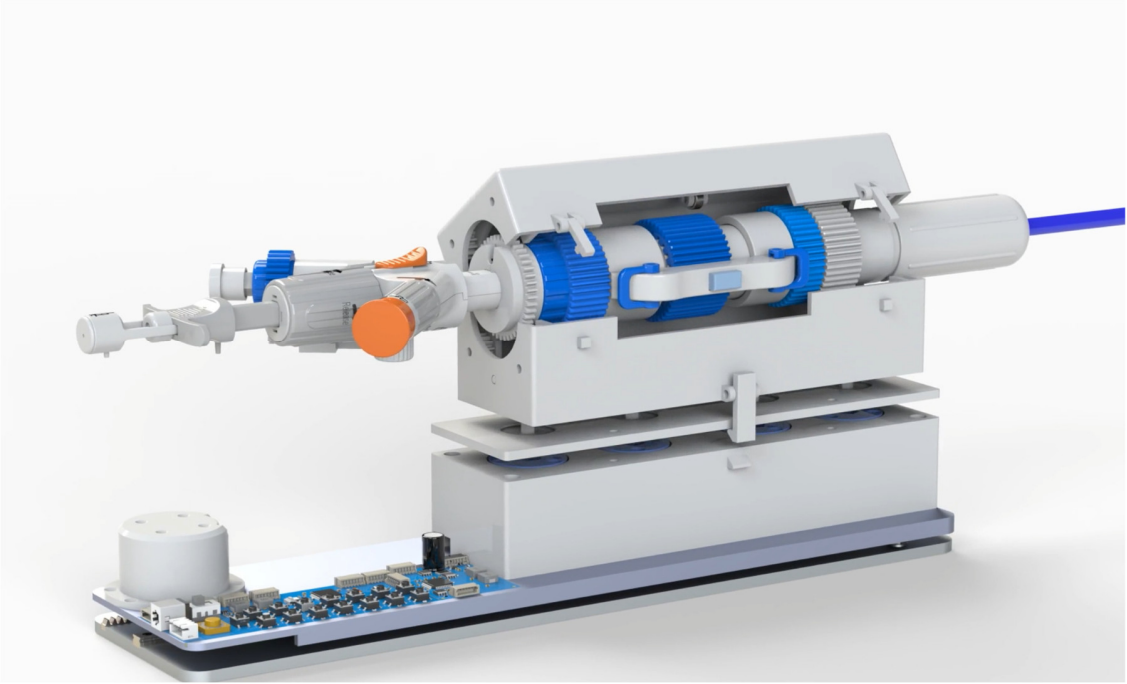

**Movie S2. Animated schematic of the detachable modular design of the proposed robotic drive and consumables (See file movie_S2 in the Supplementary Materials).**

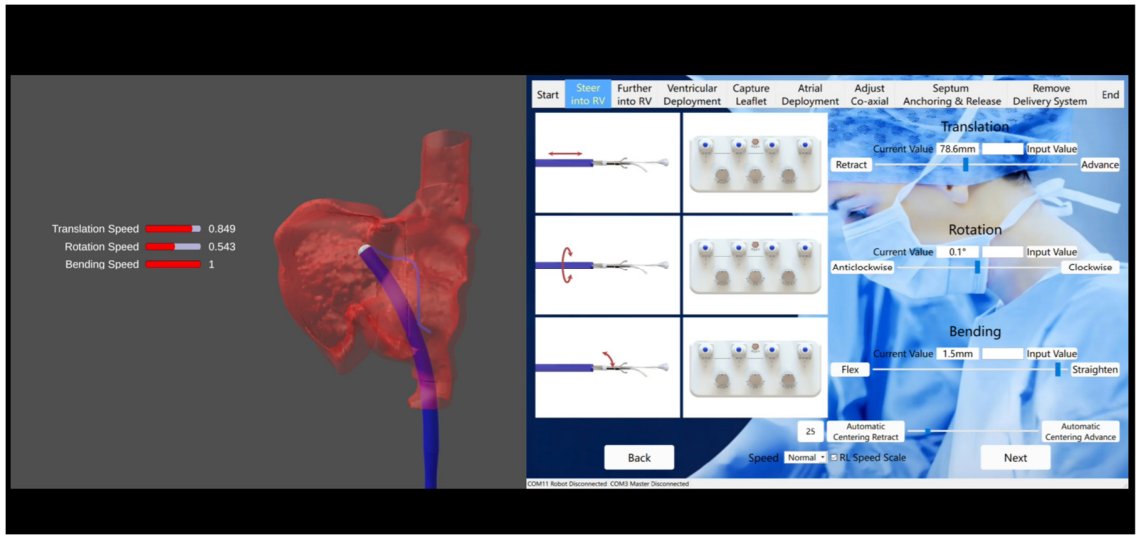

**Movie S3.** Demonstration of the autonomous localization strategy obtained through reinforcement learning in the pre-planning environment (See file movie_S3 in the Supplementary Materials).

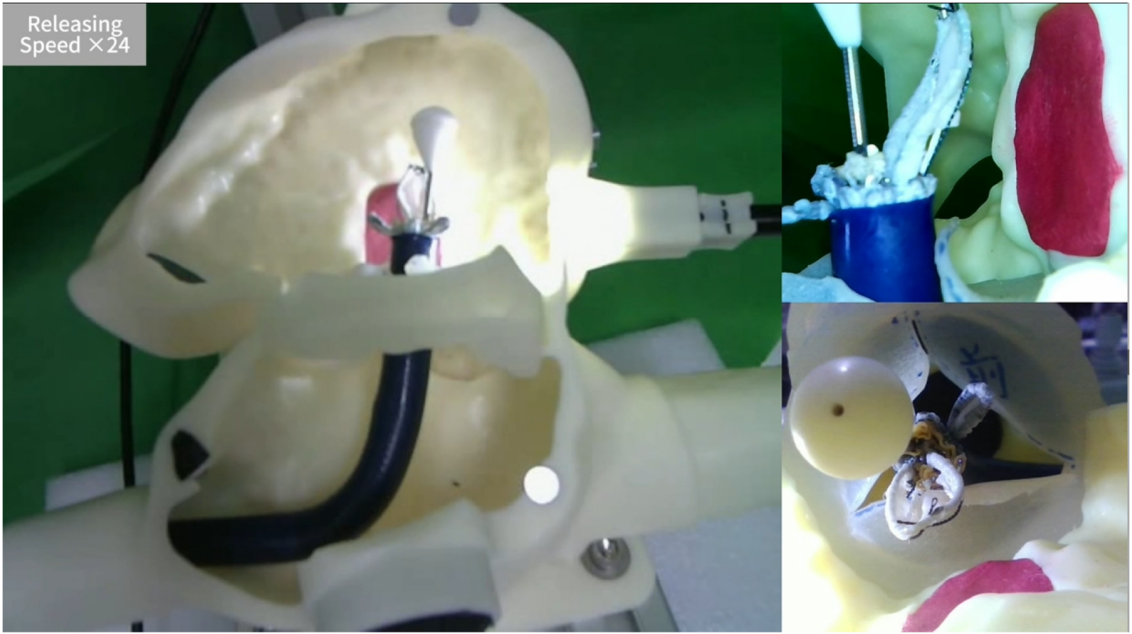

**Movie S4. Video recording of an example in-vitro phantom experiment (See file movie_S4 in the Supplementary Materials).**

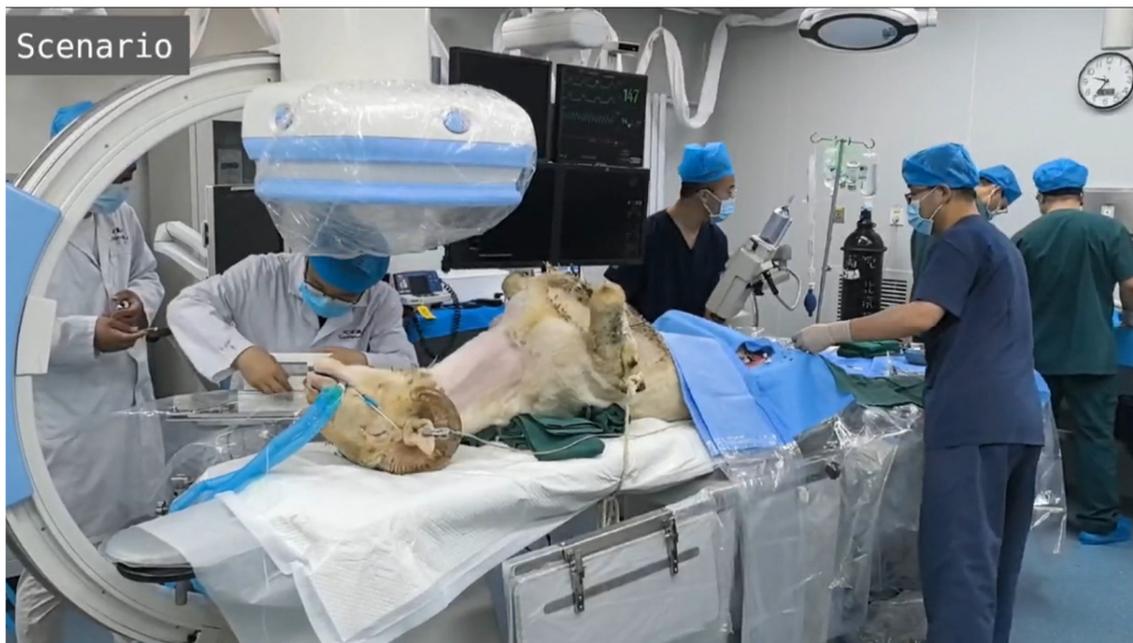

**Movie S5. Video recording of the in-vivo full-process experimental procedure (See file movie_S5 in the Supplementary Materials).**